\documentclass[10pt,conference]{IEEEtran}
\IEEEoverridecommandlockouts

\usepackage{cite}
\usepackage{amsmath,amssymb,amsfonts}
\usepackage{algorithmic}
\usepackage{graphicx}
\usepackage{times}
\usepackage{latexsym}
\usepackage[T1]{fontenc}
\usepackage[utf8]{inputenc}
\usepackage{graphicx}
\usepackage{graphicx}
\usepackage{subfigure}
\usepackage{hyperref}
\usepackage{subcaption}
\usepackage{amsmath}
\usepackage{amssymb}
\usepackage{tikz}
\usepackage{makecell}
\usetikzlibrary{fit,positioning}
\usepackage{pgfplots}
\usepgfplotslibrary{fillbetween}
\usepackage{enumitem}
\usepgfplotslibrary{groupplots}
\usepackage{booktabs}
\usepackage{algorithm}
\usepackage{algorithmic}
\usepackage{multirow}
\usepackage{inconsolata}
\usepackage{microtype}
\usepackage{textcomp}
\usepackage{tabularx}
\usepackage{multirow}
\usepackage{booktabs}
\usepackage{xcolor}
\begin{document}

\title{Label-Confidence-Aware Uncertainty Quantification in Natural Language Generation\\
}
\author{
\IEEEauthorblockN{
 Qinhong Lin\textsuperscript{a}, Yinglun Feng\textsuperscript{a}, Yuhao Zhang\textsuperscript{a}, Zhongliang Yang\textsuperscript{b,a,*}, Linna Zhou\textsuperscript{b,a,*}
}
\IEEEauthorblockA{
\textsuperscript{a}School of Cyberspace Security, Beijing University of Posts and Telecommunications, Beijing, China
}
\IEEEauthorblockA{
\textsuperscript{b}QuanCheng Laboratory, Jinan, China
}
\IEEEauthorblockA{
\{greenred99, yangzl, zhoulinna\}@bupt.edu.cn
}
}

\maketitle

\begin{abstract}
Large Language Models (LLMs) demonstrate remarkable capabilities in generative tasks but pose potential risks due to their tendency to generate hallucinatory responses. Therefore, Uncertainty Quantification (UQ), which aims to distinguish the validity of answers, is crucial for ensuring the safety and robustness of AI systems. However, existing methods primarily rely on measuring the entropy of multiple stochastic samples to represent uncertainty, often overlooking the specific uncertainty information associated with the candidate answer under evaluation. This oversight can lead to biased classification outcomes. In this paper, we investigate the discrepancy between global entropy from multiple samples and local confidence of candidate answer, and propose a Label-Confidence-Aware Uncertainty Quantification (LCA-UQ) method based on Pointwise Kullback-Leibler (PKL) divergence. Our method effectively bridges the gap between the consistency of sampled outputs and the calibration of the candidate answer, thereby enhancing the reliability and stability of uncertainty assessments. Empirical evaluations across a range of popular LLMs and NLP datasets reveal that label sources significantly impact classification. Furthermore, our approach effectively captures the nuances between sampling results and label sources, demonstrating superior performance in uncertainty estimation.
\end{abstract}

\definecolor{clr1}{RGB}{234,131,121} 
\definecolor{clr2}{RGB}{125,174,224} 
\definecolor{clr3}{RGB}{244,164,96} 
\definecolor{clr4}{RGB}{70,130,180} 
\definecolor{clr5}{RGB}{143,188,143} 
\section{Introduction}
\begin{figure*}[t]
    \centering
    \includegraphics[width=0.7\linewidth]{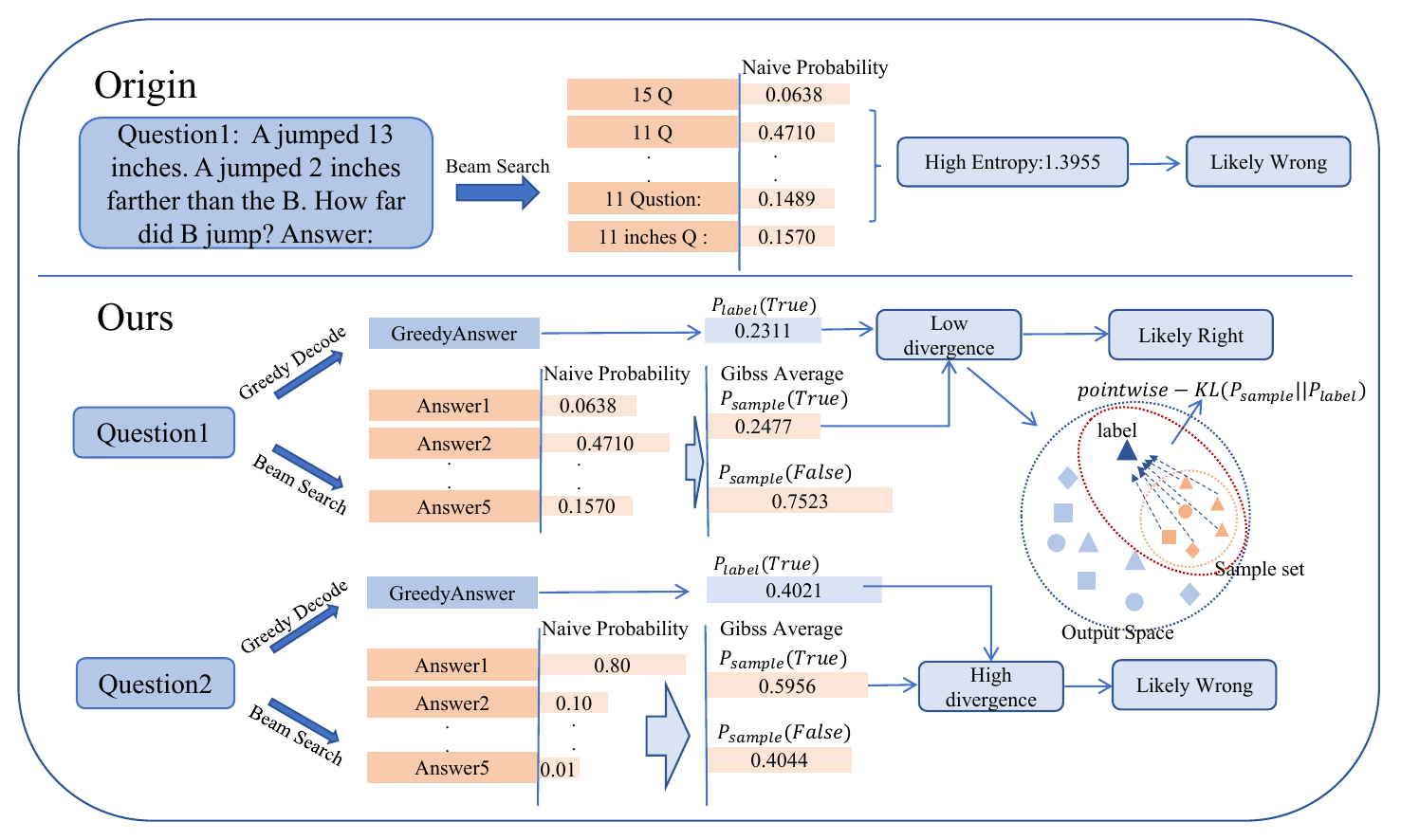} 
    \caption{Ignoring the probability information of the label answer in free-form may lead to incorrect uncertain classification. We term it as label confidence unawareness, and integrate the omitted information into our method.}
    \label{fig:framework}
\end{figure*}

Large language models (LLMs) have demonstrated formidable capabilities in natural language processing tasks such as machine translation~\cite{fomicheva2020unsupervised}, abstract text summarization~\cite{brown2020language}, and question-answering~\cite{touvron2023llama}. 
Techniques such as In-context Learning (ICL)~\cite{dong2022survey} and Chain-of-Thought (CoT)~\cite{wei2022chain} have further enhanced model performance on complex reasoning tasks and scenarios involving unseen data, consistently setting new benchmarks. 
However, despite their proficiency under scaling laws~\cite{kaplan2020scaling}, these models underperform on more challenging tasks like mathematical problems~\cite{luo2023wizardmath}. 
A significant concern is that, rather than refusing to answer, models are more likely to generate answers that include illusory reasoning processes and hallucinations. 
Uncertainty estimation and measurement have become essential tools aiding in determining the extent to which humans can trust AI-generated content.
Previous works in this field have involved prompting LLMs to self-assess the confidence of their own answers or employing confidence or entropy assessments based on model outputs. 
\textit{Semantic Entropy} (SE)~\cite{kuhn2023semantic} has introduced semantic-based entropy prediction schemes in that account for the synonym phenomena inherent in language models, performing answer aggregation in semantic space. 
\textit{SAR}~\cite{duan2023shifting} and \textit{MARS}~\cite{bakman2024mars} focus on more precisely measuring about the information content based on semantic importance weighting to offer viable approaches to align the sampling entropy more closely with the actual value.
LUQ~\cite{zhang2024luq} and CotUQ~\cite{zhang2025cot} focus on long generation scenarios, further expanding the application scope of UQ.
In existing works, single-sample-based approaches fail to capture the weight of samples within the overall sampling distribution, resulting in its high dependence on calibration. Although approaches that use consistency across multiple samples for uncertainty quantification (UQ) have shown strong results, a common issue in current research is the gap between consistent information in the sampling space and individual calibration.
As shown in Fig.~\ref{fig:framework}, when given a question, in the beam search multi-sampling strategy, three out of the five answers generated by the LLM are correct, but due to the high overall entropy value, the LLM may be marked as unable to answer this question. 
Such an error arises because the entropy threshold used in the evaluation considers only the absolute value, such as the common $-\log(0.5)$, while disregarding the model's own distribution for the question. Specifically, the greedy decoding probability is lower than the probability corresponding to the sample entropy value, which is 0.1661, as shown below.
We further quantitatively analyze this issue as well as the fragility of existing UQ methods in the Preliminary section.

To mitigate this issue, in this work, we propose a label-confidence-aware uncertainty quantification (LCA-UQ) method, as shown in Fig.~\ref{fig:framework}, based on Pointwise KL-Divergence (PKL) to bridge the gap between self-consistency information and calibration information.
Specifically, we first sample answers from the model's distribution space, as well as the output probabilities to calculate the self-consistency and entropy of the sample set. We then convert the entropy into the Gibbs probability, representing the overall probability of the model's sampling space. Next, we select a candidate answer, such as the greedy decoding answer, to represent the model's response to the question and obtain its probability. Finally, we use the Pointwise Kullback-Leibler divergence (PKL) between the two information as a measure of uncertainty, to classify whether the model can answer the question and whether the answer can be trusted.
The LCA-UQ method can be integrated with Monte Carlo and semantic clustering on any single-sample confidence evaluation scheme, establishing its flexibility.
We conducted experiments using multiple different entropy measurement methods as baselines, and the results demonstrate the robustness of LCA.
Our contributions are as follow:
\begin{itemize}
  \item We build on preliminary experiments to propose insights about bias in evaluating different subjects that further influence the process of uncertainty quantification candidate selecting and analyze the fragility of uncertainty.
  \item We introduce a novel and flexibility framework for estimating uncertainty, LCA-UQ, which utilizes pointwise KL-Divergence to explicitly account for the discrepancies between candidate answer confidence and the entropy of global samples.
  \item We evaluate multiple important free-form question-answering datasets in a CoT paradigm on popular pre-trained LLMs ranging from $4B$ to $32B$. Results demonstrate that our LCA-UQ surpass baseline methods. 
  Furthermore, through hyperparameter ablation experiments, we demonstrate how the variables in our method impact the final results and analyze its robustness.
\end{itemize}

Our data/code is available at \url{https://github.com/linqinhong/LCA}.

\section{Related Work}
Verbalization and logit-based or entropy-based methods play a crucial role in addressing uncertainty in the field of Natural Language Processing (NLP). 
The verbalization methods which prompt models to output confidence levels for their generated content, first introduced by~\cite{lin2022teaching}, unfortunately often result in overconfident outputs. Enhancements such as COT reasoning~\cite{xiong2023can} and multi-round dialogue cross models~\cite{cohen2023lm} encourage models to stimulate multi-steps reasoning for a more convincing scores.
Fine-tuning methods transforms model confidence outputs into assessments of answer correctness in a designed format and tuning the models with specially crafted data~\cite{kapoor2024calibration,han2024enhancing}.
Logit-based and entropy-based methods assess model confidence and uncertainty by focusing on the logits during the output process. Reference~\cite{kadavath2022language} add a classification head to the model's final layer, mapping logits to the probability of the “True” token, thus estimating the model's confidence in its responses. 
Reference~\cite{huang2023look} combine token-level probabilities and one-sentence entropy to evaluate the uncertainty in model-generated content. 
Reference~\cite{jiang2021can} proposes to mitigate the miscalibration of token probability caused by linguistic synonymy through data augmentation training and temperature finetuning and
~\cite{farquhar2024detecting} suggests that aggregates probabilities of synonymous sentences at the sentence-level in the multi-sampling process for better hallucination detection.

\section{Preliminary}
\noindent \textbf{Background.}
Uncertainty Quantification (UQ) aims to compute a prior metric that is highly correlated with answer correctness, used to predict the reliability of the model's output in the absence of a known reference. It can be understood as the entropy of the model's predictions, namely Predictive Entropy (PE). For a given input $x$ and output space $Y$, the predictive entropy is calculated as follows:
\begin{align}
PE(x) = -\sum_{y_i\in Y}(P(y_i|x)\log P(y_i|x))
\end{align}
where $P(y_i|x)$ is the conditional probability of a generation $y_i$.
The higher $PE(x)$ is, the closer the model's output probabilities are to a uniform distribution, indicating lower confidence in any specific output $y$ out of the output space $Y$, and thus greater model uncertainty.

In Bayesian networks, the sampling space for a model with a vocabulary of $K$ tokens generating sequences of length $L$ is exponentially large, specifically $|K|^L$, posing computational challenges. To mitigate these challenges, we can employ Monte Carlo sampling~\cite{gal2016dropout}, which introduces random factors to approximate the sampling process.
An unbiased estimate of entropy can be:
\begin{equation}
    \begin{aligned}
        PE(x) & = -\frac{1}{|N|}\sum_{y_i \in  Y} \log P(y_i|x) 
    \end{aligned}
\label{equ:Entropy_approximate}
\end{equation}

As probabilities tend to decrease with increasing length, length-normalization method~\cite{malinin2020uncertainty}, replacing probability of $y$ with $\frac{1}{N}\sum_{i}^{N}\log P(y_i|y_{<i})$, could be used to scale the conditional probabilities of sentences of different lengths to the same magnitude and has been successfully applied in machine translation scenarios~\cite{murray2018correcting}. 



\noindent \textbf{Bias in Evaluating Different Objects.}
Previous uncertainty workflows typically involve first sampling a candidate answer through greedy decoding and obtaining probabilities, then sampling $N$ samples to measure entropy in order to estimate the quality of the greedy decoded answer. However, they overlook an important issue: when the number of samples is limited, the sampling space may fail to cover the candidate answer adequately.
To analyze the representativeness of the greedy decoded label, we evaluated the relationship between candidate result and the sampled results. We categorize the test question samples based on whether the semantic meaning of the candidate answer is covered by the sample set ("In" vs. "Not in"). 
Specifically, we first measured the NLI score between the candidate and every samples. Denoting sample set as $\mathcal{S}$ and the greedy decoding answer as $g$, $g$ is considered to be cover by $\mathcal{S}$ if at least one $NLI(\mathcal{S}_i, \mathcal{g})$ exceeds a predefined threshold $\alpha$:
\begin{align}
\text{sim}(\mathcal{S},\mathcal{G}) = \begin{cases}1 & if\ \exists~\text{NLI}(\mathcal{S}_i,\mathcal{G})>\alpha \\0&\text{otherwise} \end{cases}
\end{align}
We set $\alpha$ to 0.5, consistent with prior works.
\begin{figure}
    \subfigure[2WikimultihopQA]{
        \centering
        \resizebox{0.23\textwidth}{!}{
        \begin{tikzpicture}[trim axis right, baseline]
        \begin{axis}[
            ybar stacked,
            xtick=data,          
            tick align=inside,
            legend style={at={(0,1)},anchor=south west,legend columns=-1, draw opacity = 0.5},
            xticklabels={ 
            {\shortstack{Llama3.1\\8B}}, 
            {\shortstack{Qwen3\\4B}},
            {\shortstack{Qwen3\\14B}},
            {\shortstack{Qwen3\\32B}}},  
            xlabel={Data\& Num},  
            ylabel={Percentage},      
            ylabel style={yshift=-10pt,font = \LARGE},
            xlabel style={yshift=-10pt,font = \LARGE},
            ymin=0,               
            ymax=1,               
        ]
        \addplot[draw=clr3,fill=clr3]           
        coordinates
        {
         (1,0.43)
         (2,0.3033)
         (3,0.3166)
         (4,0.37)
        };
        \addlegendentry{not in sample}
        \addplot[draw=clr4,fill=clr4]         
        coordinates
        {
        (1,0.58)
        (2,0.70)
        (3,0.69)
        (4,0.63)
        };
        \addlegendentry{in sample}
        \end{axis}
        \end{tikzpicture}
            }
}
    \hspace{-0.02\textwidth}
    \subfigure[HotpotQA]{
        \centering
        \resizebox{0.23\textwidth}{!}{
        \begin{tikzpicture}[trim axis right, baseline]
        \begin{axis}[
            ybar stacked,
            xtick=data,          
            tick align=inside,
            legend style={at={(0,1)},anchor=south west,legend columns=-1, draw opacity = 0.5},
            xticklabels={ 
            {\shortstack{Llama3.1\\8B}},%
            {\shortstack{Qwen3\\4B}},%
            {\shortstack{Qwen3\\14B}},%
            {\shortstack{Qwen3\\32B}}}, 
            xlabel={Data\& Num},  
            ylabel={Percentage},      
            ylabel style={yshift=-10pt,font = \LARGE},
            xlabel style={yshift=-10pt,font = \LARGE},
            ymin=0,               
            ymax=1,               
        ]
        \addplot[draw=clr3,fill=clr3]           
        coordinates
        {
         (1,0.3866)
         (2,0.3033)
         (3,0.3067)
         (4,0.3)
        };
        \addlegendentry{not in sample}
        \addplot[draw=clr4,fill=clr4]         
        coordinates
        {
        (1,0.62)
        (2,0.70)
        (3,0.70)
        (4,0.7)
        };
        \addlegendentry{in sample}
        \end{axis}
        \end{tikzpicture}
            }
        }
    \caption{Percentage of 2WikimultihopQA w. \& w/o label answers in sample on four models. 'in sample' represents the greedy decoded answer can be entailed by at least one of the sampled answers, while 'not in sample' means it is not entailed by any of the sampled answers.}
    \label{fig:greedy_in_vs_not_in}
\end{figure}
Fig.~\ref{fig:greedy_in_vs_not_in} illustrates the occurrence of greedy decoding results within the sampled outcomes for four models over 2WikimultihopQA~\cite{ho2020constructing} and HotpotQA~\cite{yang2018hotpotqa}. 
Our results indicate that in many cases, the candidates do not appear within the sampled set.
We further grouped the test data according to whether it is in or not in sample set to analyze the impact on the classification performance of the set entropy. 
We conduct experiments on these two datasets using three models and two methods. We present the experimental results in Tab.~\ref{tab:AUROC_with_without}. 
In scenarios "In", standard uncertainty metrics exhibit a strong correlation with correctness, yielding high AUROC scores. Conversely, in "Greedy-Sample Mismatch" scenarios ("not in"), we observe a precipitous decline in the discriminative power of the baseline.
This phenomenon indicates that the reliability of existing uncertainty workflows is highly contingent on the sampling coverage. 
Therefore, we emphasize that an uncertainty measurement scheme based on entropy must consider whether the candidate answer is covered by the sampled instances. The simplest yet efficient method is to integrate the candidate into the sample for calculating entropy. In the experiments of this paper, we ensure this condition is strictly met.

\begin{table}[]
\caption{Uncertainty estimation AUROC results of \textit{LNPE} \& \textit{LUQ-Pair} with and without candidate answers in sample set. }
\centering
\begin{tabular}{c|c|cc|cc}
\toprule
\multirow{2}{*}{Dataset}         & \multirow{2}{*}{Model} & \multicolumn{2}{c}{LNPE} & \multicolumn{2}{c}{LUQ-Pair} \\
                                 &                        & in         & not in      & in            & not in        \\
\midrule
\multirow{3}{*}{2WikimultihopQA} & Llama-3.1-8B           & 72.25      & 39.57       & 71.52         & 42.01         \\
                                 & Qwen3-14B              & 68.80      & 68.82       & 76.55         & 72.68         \\
                                 & Qwen3-4B               & 63.12      & 66.11       & 80.29         & 50.76         \\
\midrule
\multirow{3}{*}{HotpotQA}        & Llama-3.1-8B           & 81.76      & 42.60       & 85.72         & 53.47         \\
                                 & Qwen3-14B              & 64.81      & 61.59       & 76.55         & 72.68         \\
                                 & Qwen3-4B               & 60.70      & 40.53       & 80.29         & 50.76        \\
\bottomrule
\end{tabular}
\label{tab:AUROC_with_without}
\end{table}

\begin{figure}[htbp]
    \centering
    \label{fig:fragile}
    \includegraphics[width=0.7 \linewidth]{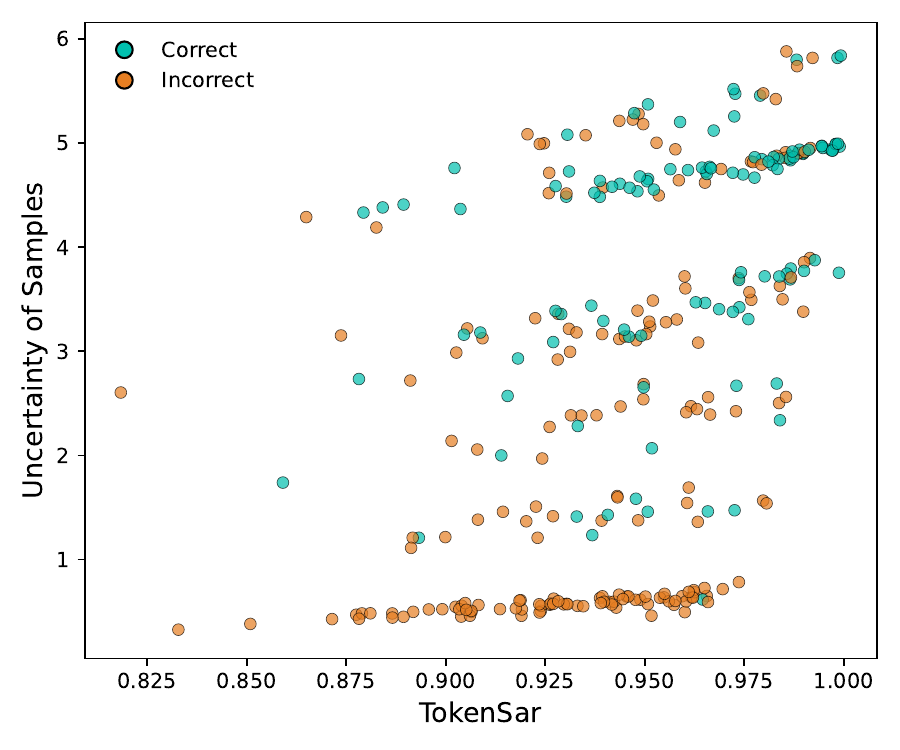} 
    \caption{Visualization of the relationship between sample uncertainty and candidate probability on 2WikimultihopQA. The scattered distribution indicates that relying solely on one dimension is fragile for capturing complex error patterns.}
    \label{fig:plot}
\end{figure}

\noindent \textbf{Fragility of Uncertainty}
In the previous section, we provided an example to illustrate the potential fragility when consistency-based and confidence-based methods work in isolation and we conduct a deeper analysis here. Fig.~\ref{fig:plot} illustrates the relationship between sample-based uncertainty and confidence-based uncertainty for Qwen3-4B~\cite{yang2025qwen3} on the 2WikimultihopQA dataset, distinguishing between correct and incorrect predictions. For the confidence-based method, we selected TokenSAR~\cite{duan2023shifting}, while for the entropy-based method, we compute semantic clustering entropy based on TokenSAR.
We observe a complex interaction between the two metrics: 1. In low-uncertainty regimes, the model output is consistent, yet the probability score serves as a critical discriminator. Specifically, samples with lower probabilities in this region strongly tend to be incorrect, indicating that consistency without confidence is often a sign of error. 2.In high-uncertainty regimes, despite the dispersion in the semantic space, we observe a significant number of correct answers even with moderate probabilities. These patterns highlight the fragility of relying solely on either uncertainty or probability. The misalignment between high semantic dispersion and local confidence suggests that neither metric alone is sufficient, motivating our proposed fusion approach.

\section{Method}
In this section, we present the implementation process of the LCA-UQ framework, which includes: (1) candidate answer sampling, (2) distribution space estimation, and (3) bridging local calibration and global consistency.

\noindent\textbf{Candidate Answer Sampling.}
Given a query $x$ and model $\mathcal M$ we first generate a candidate answer $g$ via greedy decoding. In order to explore the model's output distribution, we then employed multinomial sampling to generate |$S'$| sequences, forming the sample set $S'=\{s_1, s_2, ..., s_m\}$.
Crucially, to ensure the candidate answer is covered by the semantic space, we construct an augmented set $S=\{s_0, s_1, ..., s_m\}$ where $s_0$ is the greedy hypothesis $g$.
We then apply a confidence estimation framework to quantify the reliability $C(s_i|x)$ of each prediction. While our method is agnostic to the underlying uncertainty backbone, it can also be calculated from entropy-based metrics such as average token entropy. Let $E(s_i|x)$ denote the measured entropy, we derive the final confidence score $C(s_i|x)$ via the exponential transformation:
\begin{equation}
 C(s_i|x) = \exp(-E(s_i|x)).   
\end{equation}
 
Through $C(g|x)$, we can measure the model's local calibration for the question $x$.
Here, the forced insertion of $s_0$ into the sample set is a widely used configuration in prior works. Furthermore, as demonstrated in Tab~\ref{tab:AUROC_with_without}, this setting is essential.

\noindent\textbf{Distribution space estimation.}
Through Monte Carlo sampling, we can use the sampled instances to unbiasedly estimate the distribution space predictive entropy of model $\mathcal{M}$ for the question as shown in Eq.~\ref{equ:Entropy_approximate}.
In natural language generation tasks, different sentences may express the same meaning, thus sharing a common semantic space. Considering that the correctness of a non-multiple-choice answer should be judged based on its semantics rather than just its lexical content, we calculate the entropy of the sample space after semantic clustering of multiple samples, rather than using the lexical-independent sample entropy.
We computed the relevant score $P(entail|s_i,s_j)$ between each answer using a pretrained language model as follow:
\begin{equation}
    P(entail|s_i,s_j) = \frac{\exp(l_e)}{\exp(l_e)+\exp(l_c)},
\end{equation}
where $l_e$ and $l_c$ are the logits of the "entailment" and "contradiction" classes respectively. We treat samples with a neutral relationship as irrelevant samples as well. Specifically, we used RoBERTa-Large~\cite{liu2019roberta} for calculating.
Samples with bidirectional semantic entailment will be clustered together, and these samples will ultimately be categorized into |$\mathbb C$| clusters.
The conditional probability of a cluster is the sum of the probabilities of the sentences, and we can calculate $E(S|x)$ with Monte Carlo Approximation, the entropy in the semantic space as follows:
\begin{equation}
\begin{aligned}
    E(S|x) &=-\sum_{c \in \mathbb C}C(c_i|x)\log C(c_i|x)\\ &= -\sum_{c \in \mathbb C}((\sum_{s_i\in c}C(s_i|x)\log (\sum_{s_i\in c}C(s_i|x))))\\ & \approx -|\mathbb{C}|^{-1}\sum_{i=1}^{|\mathbb C|}\log C(c_i|x).
\end{aligned}
\end{equation}
This entropy encapsulates the global consistency information as well as uncertainty of the model's sampling space.
We then calculate the Gibbs probability $C(S|x) = \exp(-E(S|x))$ as the confidence of S, representing the model's perceived probability of a set to be able to provide an answer. 

\noindent\textbf{Bridging local calibration and global consistency.}
We reframe uncertainty estimation as a measurement of divergence between the greedy decoding confidence $P$ (approximated as a point mass) and the aggregated semantic distribution $Q$ derived from sampling.
Specifically, to quantify the information loss—or surprisal—incurred when the greedy answer is evaluated against the semantic consensus, we employ the Pointwise Kullback-Leibler divergence (PKL) defined as follow:
\begin{equation}
    U(S|x) =P\log \frac{P}{Q} =  C(g|x) \cdot \log \frac{C(g|x)}{C(S|x)} .
\end{equation}
This formulation inherently resolves the limitations identified in our preliminary analysis:
1.confidence Anchoring: the term $C(g|x)$ integrates the model's intrinsic local confidence; 2. mismatch Penalty: as the log-ratio term diverges, it imposes a severe penalty on the uncertainty score.


\begin{table*}[]
\caption{Uncertainty estimation AUROCs of our LCA method with different methods as backbone and baselines across datasets. Bold indicates better performance.}
\renewcommand{\arraystretch}{1} 
\centering
\begin{tabular}{c|l|cc|cc|cc|cc|cc}
\toprule
\multirow{2}{*}{model}        & \multirow{2}{*}{dataset} & \multicolumn{2}{c|}{LNPE} & \multicolumn{2}{c|}{TokenSAR} & \multicolumn{2}{c|}{Cot-SE} & \multicolumn{2}{c|}{LUQ} & \multicolumn{2}{c}{LUQ-PAIR} \\
                              &                          & Base        & LCA        & Base          & LCA          & Base         & LCA         & Base       & LCA        & Base          & LCA           \\
\midrule
\multirow{4}{*}{Llama-3.1-8B} & 2WikimultihopQA    & 72.96       & \textbf{75.04}      & 72.96  & \textbf{86.19}  & 74.79  & \textbf{87.15} & 67.10 & \textbf{86.37} & 64.48 &\textbf{ 84.04}           \\
                              & HotpotQA           & 77.76       & \textbf{78.92}      & 77.76  & \textbf{86.16}  & 75.22  & \textbf{86.97} & 82.51 & \textbf{85.61} & 81.18 &\textbf{ 85.51}           \\
                              & MedQA              & 67.41       & \textbf{69.74}      & 67.41  & \textbf{82.31}  & 64.98  & \textbf{82.31} & 77.32 & \textbf{81.43} & 76.52 &\textbf{ 82.25}          \\
                              & Math               & 78.05       & \textbf{82.01}      & 78.05  & \textbf{87.14}  & 79.67  & \textbf{87.75} & 88.81 & \textbf{89.60} & \textbf{90.37} &\textbf{ }89.11         \\
\midrule
\multirow{4}{*}{Qwen3-4B}     & 2WikimultihopQA    & 68.25       & \textbf{80.15}      & 68.23  & \textbf{80.15}  & 73.37  & \textbf{81.28} & 77.26 & \textbf{81.80} & 77.11 &\textbf{ 82.29}           \\
                              & HotpotQA           & 67.98       & \textbf{86.36}      & 67.99  & \textbf{86.37}  & 63.43  & \textbf{86.64} & 83.15 & \textbf{87.04} & 84.11 &\textbf{ 86.14}           \\
                              & MedQA              & 62.33       & \textbf{67.20}      & 62.34  & \textbf{67.08}  & 58.81  & \textbf{66.09} & 67.40 & \textbf{69.73} & 65.08 &\textbf{ 69.76}           \\
                              & Math               & 85.78       & \textbf{90.57}      & 85.79  & \textbf{90.57}  & 84.69  & \textbf{91.71} & 89.44 & \textbf{96.58} & 81.27 &\textbf{ 96.29}           \\
\midrule
\multirow{4}{*}{Qwen3-14B}    & 2WikimultihopQA    & 78.39       & \textbf{82.11}      & 78.38  & \textbf{82.12}  & 78.81  & \textbf{82.74} & 72.24 & \textbf{82.85} & 73.57 &\textbf{ 81.98 }          \\
                              & HotpotQA           & 67.48       & \textbf{81.74}      & 67.47  & \textbf{81.27}  & 63.16  & \textbf{80.35} & 81.33 & \textbf{81.96} & 79.66 &\textbf{ 83.17 }         \\
                              & MedQA              & 70.09       & \textbf{76.70}      & 70.09  & \textbf{77.18}  & 66.14  & \textbf{74.73} & 67.04 & \textbf{69.91} & 65.59 &\textbf{ 70.46 }          \\
                              & Math               & 83.00       & \textbf{95.06}      & 83.02  & \textbf{95.04}  & 88.01  & \textbf{95.30} & 92.32 & \textbf{98.76} & 90.01 &\textbf{ 98.68 }          \\
\midrule
\multirow{4}{*}{Qwen3-32B}    & 2WikimultihopQA    & 72.19       & \textbf{83.48}      & 72.20  & \textbf{84.35}  & 73.23  & \textbf{84.22} & \textbf{82.60} & 81.96 &\textbf{ 83.35}  &79.36           \\
                              & HotpotQA           & 68.74       & \textbf{87.78}      & 68.75  & \textbf{87.62}  & 66.23  & \textbf{87.31} & 84.23 & \textbf{85.40} & 84.20 &\textbf{ 84.70}          \\
                              & MedQA              & 71.77       & \textbf{75.69}      & 71.77  & \textbf{86.12}  & 67.77  & \textbf{85.14} & 78.58 & \textbf{83.99} & 78.83 &\textbf{ 80.38}         \\
                              & Math               & 82.53       & \textbf{96.74}      & 82.49  & \textbf{97.54}  & 83.18  & \textbf{97.64} & 85.93 & \textbf{95.82} & 87.79 &\textbf{ 94.33}         \\
\midrule
\multicolumn{2}{c|}{Average} & 73.42 & \textbf{81.08} & 73.42 &\textbf{84.83} &72.59 &\textbf{84.83} &79.83 &\textbf{84.93} &78.94 &\textbf{84.28} \\
\bottomrule
\end{tabular}
\label{tab:AUROC_of_baseline_VS_KLD}
\end{table*}

\section{Experiments}
\label{sec:experiments}

\noindent\textbf{Datasets.} 
As the capabilities of LLMs have improved, the scenarios for evaluating UQ methods in short-generation tasks are limited. Therefore, we focus our experimental scenarios on long-generation tasks, CoT specifically. We require models to first generate a chain-of-thought for reasoning and then provide a final answer. We conduct experiments on several free-form text generation tasks in NLP, including two multi-hop question-answering datasets, 2WikimultihopQA~\cite{ho2020constructing} and HotpotQA~\cite{yang2018hotpotqa}, a domain-specific question-answering dataset, MedQA~\cite{jin2021disease}, and a mathematical reasoning dataset, Math~\cite{hendrycks2021measuring}.

\noindent\textbf{Baselines.}
We selected two classic UQ methods: Length Normalizaiton Predictive Entropy (LNPE)~\cite{malinin2020uncertainty}, TokenSAR~\cite{duan2023shifting}, and extended them to long-generation tasks. Additionally, we included three state-of-the-art uncertainty quantification methods work well in long-generation: CoT-UQ~\cite{zhang2025cot}, LUQ and LUQ-Pair~\cite{zhang2024luq} as the backbone and compared the performance of these methods before and after applying LCA. 
For CoT-UQ, it is a single-sample confidence evaluation method, and we extend it by combining it with SE. We label it as CoT-SE in the following sections.
For other these methods, candidate probabilities are calculated according to the approach outlined in the respective papers. When performing semantic clustering, considering the semantic richness of long-generation answers, directly aggregating them is not sufficiently accurate. Therefore, we disregard CoT and only use the final model output for NLI calculation and classification.

\noindent\textbf{Models.}
The uncertainty measurement methods we selected require token probabilities. We conducted experiments using popular open-source LLMs, including Llama-3.1-8B~\cite{dubey2024llama3} and Qwen3-4/14/32B~\cite{yang2025qwen3}, covering models of different sizes.

\noindent\textbf{Metric.}
For correctness assessment, we leveraged a majority voting ensemble of three closed-source APIs including GPT-4.1-mini\footnote{\url{https://openai.com/index/gpt-4-1/}}, DeepSeek-V3.2-Fast\footnote{\url{https://chat.deepseek.com/}}, and Grok-4-1-Fast-Reasoning\footnote{\url{https://x.ai/grok}} to adjudicate response correctness, guaranteeing robust labeling for candidate answers. The reliability of this automated protocol is substantiated in the Appendix, where we demonstrate a high degree of consistency between the LLM voting results and human annotations.
For uncertainty method evaluation metric, we adopt Area Under the Receiver Operating Characteristic Curve (AUROC) as the primary metric to assess the efficacy of our uncertainty quantification. This metric serves as a proxy for discriminative power, measuring how reliably the uncertainty scores can separate correct generations from incorrect ones.

\noindent\textbf{Hyperparameters.}
 In the experiments, the top-P and top\-k settings for model-generated answers were set to the model's default configurations, and diversity in sampling results was ensured through polynomial sampling . For each dataset, we generated five answers per question. All experiments were carried out using two NVIDIA A40 GPUs.

\section{Results Analysis}
In Tab.~\ref{tab:AUROC_of_baseline_VS_KLD}, we provided a detailed performance comparison between the basic baselines and baselines with the LCA-UQ method on evaluation datasets. In the majority of cases, LCA-UQ effectively enhances the performance of the backbone method, demonstrating that PKL successfully links local confidence and global entropy, helping us further distinguish the correctness of answers. The most significant result of LCA-UQ on the test data is observed in the experiment with Llama-3.1-8B on the 2WikimultihopQA dataset, where the AUROC increased from 64.48\% to 84.04\%, with a 19.56\% improvement. On average, LCA-UQ showed the most notable improvement on CoT-SE, with an increase from 72.59\% to 84.83\%, an improvement of 12.24\%. In experiments with Qwen3-14B, even though the LUQ and LUQ-Pair methods achieved AUROCs of 92.32\% and 90.01\% on the Math dataset, the LCA method can further improve the results to 98.76\% and 98.68\%, respectively.

\section{Ablation Study}
\begin{figure}[t]
    \centering
    \includegraphics[width=\linewidth,height=4cm]{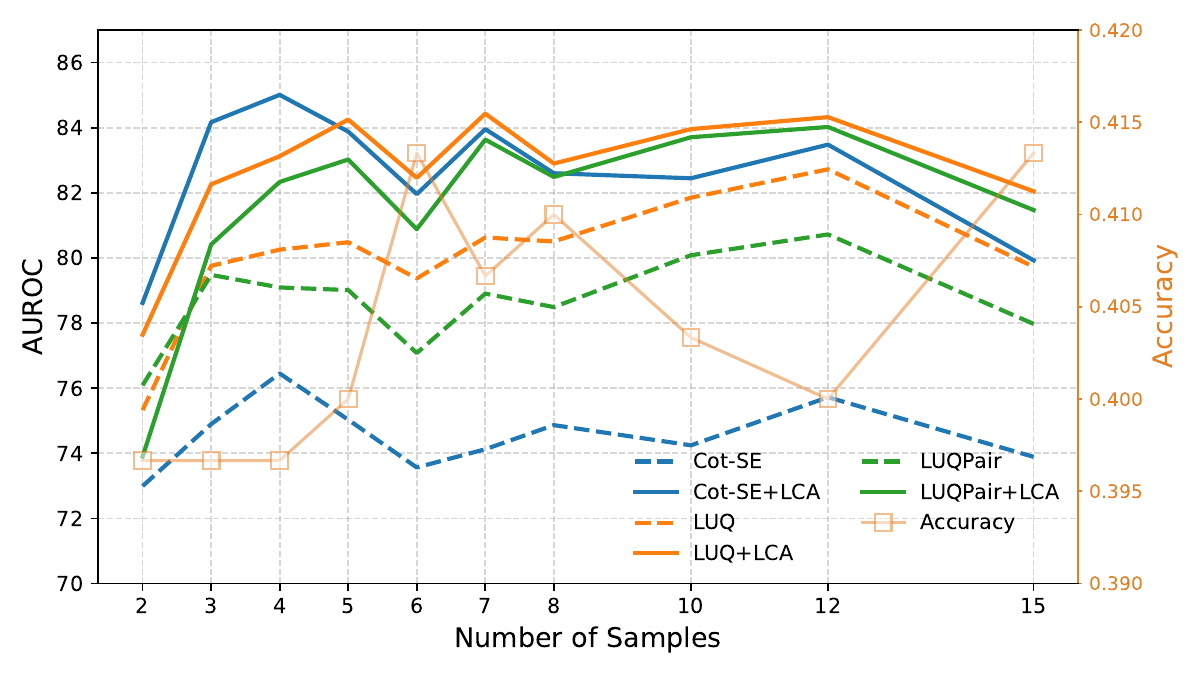} 
    \caption{AUROC versus the number of samples on HotpotQA (Llama-3.1-8B, $T=1.0$). LCA-enhanced methods (solid) consistently outperform their baselines (dashed), while task accuracy is shown on the secondary axis.}
    \label{fig:multi_sampling_auroc}
\end{figure}

\noindent \textbf{Number of Generation.}
Fig.~\ref{fig:multi_sampling_auroc} reports AUROC versus the number of samples on HotpotQA with Llama-3.1-8B ($T{=}1.0$). LCA-UQ methods (solid) consistently outperformcorresponding backbone counterparts (dashed), with the largest gains observed in the low-sample regime (2--5 samples). AUROC increases rapidly as $k$ grows and peaks around 7--12 samples, after which the improvement plateaus or slightly declines, indicating diminishing returns from additional sampling. Task accuracy, shown on the secondary axis, varies more mildly across $k$, further demonstrating that LCA-UQ effectively enhances the discriminative power of uncertainty estimation methods across different sampling budgets.
\begin{figure}[t]
    \centering
    \includegraphics[width=\linewidth,height=4cm]{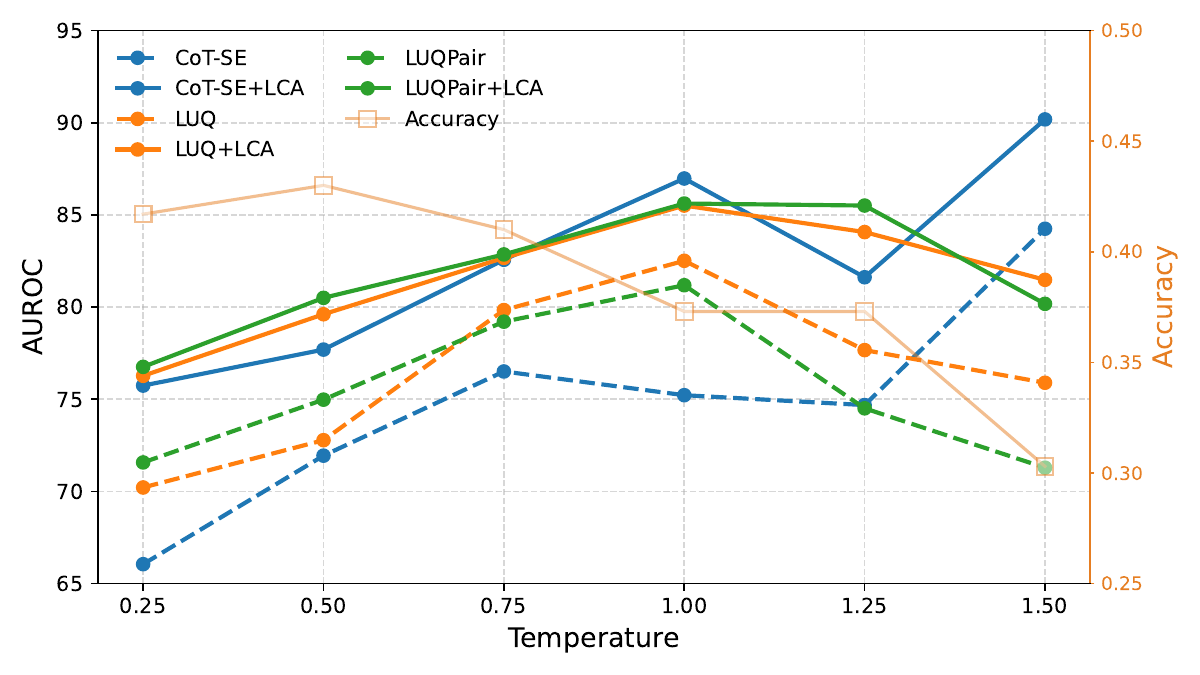} 
    \caption{Effect of temperature on AUROC for uncertainty estimation on HotpotQA. LCA (solid) improves performance over baselines (dashed) across all temperature settings; task accuracy is reported on the right axis.}
    \label{fig:ablation_result_of_Temperature}
\end{figure}

\noindent \textbf{Temperature.}
We conducted a temperature ablation experiment using Llama-3.1-8B on HotpotQA to analyze the impact of sampling temperature on AUROC.
We show the effect of temperature on performance in Fig.~\ref{fig:ablation_result_of_Temperature}.
A lower temperature results in sharper token probabilities, which reduces the diversity of the model's generation. As the temperature increases, the diversity and performance gradually improve. However, when the temperature exceeds 1, the generation quality of the LLM begins to decline, and the AUROC of both LUQ and LUQ-Pair decreases. In contrast, CoT-SE shows an increase. We attribute this to the filtering of low-importance words when measuring the entropy of individual samples in the CoT-UQ method. The entropy estimation is less affected by temperature.

\begin{table*}[htbp]
\caption{The performance of PKL-based, Mean-Based, and R-PKL-based method with three backbone methods on two datasets using three models. }
\centering
\begin{tabular}{c|c|cccc|cccc|cccc}
\toprule
\multirow{2}{*}{model}        & \multirow{2}{*}{dataset} & \multicolumn{4}{c}{CoT-SE}    & \multicolumn{4}{c}{LUQ}       & \multicolumn{4}{c}{LUQ-Pair} \\
                              &                          & Base  & Mean   & PKL   & R-PKL & Base  & ADD   & PKL   & R-PKL & Base  & Mean   & PKL   & R-PKL \\
\midrule
\multirow{2}{*}{Llama-3.1-8B} & 2Wiki                    & 74.79 & \textbf{87.13} & 87.15 & 82.73 & 67.10 & 75.76 & \textbf{84.04} & 69.63 & 64.48 & 73.01 & \textbf{86.37} & 67.10 \\
                              & HotpotQA                 & 75.22 & 79.73 & \textbf{86.97} & 64.07 & 82.51 & 85.85 & \textbf{85.51} & 73.63 & 81.18 & 83.21 & \textbf{85.61} & 69.41 \\
\midrule
\multirow{2}{*}{Qwen3-4B}     & 2Wiki                    & 78.81 & 82.71 & \textbf{82.74} & 82.51 & 72.24 & 78.50 & \textbf{81.98} & 69.57 & 73.57 & 79.77 & \textbf{82.85} & 71.97 \\
                              & HotpotQA                 & 63.16 & 79.57 & 80.35 & \textbf{81.21} & 81.33 & \textbf{83.49} & 83.17 & 74.72 & 79.66 & \textbf{81.98} & 81.96 & 73.82 \\
\midrule
\multirow{2}{*}{Qwen3-14B}    & 2Wiki                    & 73.37 & 81.08 & \textbf{81.28} & 80.53 & 77.26 & 80.02 & \textbf{82.29} & 70.58 & 77.11 & 80.57 & \textbf{81.80} & 71.49 \\
                              & HotpotQA                 & 63.43 & 86.14 & \textbf{86.64} & 86.13 & 83.15 & \textbf{86.74} & 86.14 & 81.16 & 84.11 & 87.02 & \textbf{87.04} & 82.77 \\
\bottomrule
\end{tabular}
\label{tab:AUROC_KLD_VS_R-KLD}
\end{table*}

\noindent \textbf{Different Integrate Methods}
One of the core aspects of our work is to quantify uncertainty by effectively measuring the discrepancy between the sampling distribution and the candidate answer distribution. We compare the use of pointswise KL-divergence (PKL) with the trivial method that calculate the mean value of $C(x)$ and $C(S)$ (Mean) and the method Reverse Pointwise KL-divergence (R-PKL)~\cite{malinin2019reverse} as aggregation methods, where:
\begin{equation}
    \text{U}_{R-KLD}(x)=C(S|x) \cdot \log \frac{C(S|x)}{C(g|x)}.
\end{equation}
As shown in Tab.~\ref{tab:AUROC_KLD_VS_R-KLD}, in most cases, the PKL method outperforms R-PKL and Mean.
It indicate that when we treat the sampling results as the “correct” distribution and view greedy sampling as the prediction, divergence calculations help us better identify when the model is more likely to be able to answer. 

\noindent\textbf{Effectiveness on Multi-fact Generation}
Multi-fact generation tasks represent a common category within natural language generation (NLG). We took summarization task as a representative. We utilized the Llama-3.1-8B model to conduct experiments on the XSum~\cite{narayan2018don} dataset.
Generations with ROUGE-L greater than the threshold will be assigned a label of 1, otherwise it will be assigned a label of 0. The results of these experiments are presented in Tab.~\ref{tab:xsum}.
Our LCA consistently enhances performance across various methods, achieving a maximum improvement of 0.09 on LNPE backbone. Notably, the method of LNPE performs the best. We attribute this to the presence of multiple facts in the generated text. Specifically, sequence-level clustering employed by other semantic-level methods tends to overlook the independence of individual facts within generations.

\begin{table}[htbp]
\renewcommand{\arraystretch}{1} 
\caption{Results of Llama3.1-8B on Xsum under different Rouge Threshold.}
\setlength\tabcolsep{3pt}
\centering
\begin{tabular}{c|rr|rr}
\toprule
\multirow{2}{*}{\textbf{Thres}} & \multicolumn{2}{c}{\textbf{LNPE}} & \multicolumn{2}{c}{\textbf{TokenSAR}} \\
 & base & LCA & base & LCA \\
\midrule
0.3 & 0.543 & \textbf{0.614} & 0.529 & \textbf{0.557} \\
0.2 & 0.525 & \textbf{0.616} & 0.500 & \textbf{0.532} \\
0.15 & 0.548 & \textbf{0.636} & 0.518 & \textbf{0.552} \\
\bottomrule
\end{tabular}
\label{tab:xsum}
\end{table}

\section{Conclusion}
In this paper, we reveal the impact of biases between label sources and samples in uncertainty estimation and propose our LCA method to aggregate the confidence of them. Results demonstrate that our method surpasses the state-of-the-art performance. Further ablation results show the impact of various parameters on method performance.

\section{Limitations}
We recognize that there are several areas where our approach can be further enhanced:
(1) Model Capability: Employing a more powerful model, or fine-tuning Roberta on the test domain to assess semantic relevance could yield superior sampling results for semantic clustering and would significantly boost the performance of our uncertainty measurement.
(2) Similarity Calculation in Multi-Fact Scenarios: Experiments on the Xsum dataset demonstrate that poor sequence-level similarity calculations can undermine the performance in multi-fact contexts. Implementing more refined similarity calculations in these scenarios is likely to improve overall model performance.

\section*{Acknowledgement}
This work was supported in part by Research Project of Quan Cheng Laboratory, China (Grant No.QCL20250203) and in part by the National Natural Science Foundation of China under Grant 62172053 and Grant 62302059.

\bibliographystyle{IEEEtran}
\bibliography{custom}

\appendix
\begin{table*}[htbp]
\centering
\caption{Result with selecting the answer with max likelihood inside max semanticn cluster.}
\label{tab:candidate}
\begin{tabular}{c|c|cc|cc|cc|cc|cc}
\toprule
\multirow{2}{*}{model}                 & \multirow{2}{*}{dataset} & \multicolumn{2}{c|}{LNPE} & \multicolumn{2}{c|}{TokenSAR} & \multicolumn{2}{c|}{CoT-SE} & \multicolumn{2}{c|}{LUQ} & \multicolumn{2}{c}{LUQ-Pair} \\
                                       &                          & Base        & LCA        & Base          & LCA          & Base         & LCA         & Base       & LCA        & Base          & LCA          \\
\midrule
\multirow{4}{*}{Llama-3.1-8B-Instruct} & 2Wiki          & 72.96       & 75.04      & 72.96         & 86.19        & 74.79        & 87.15       & 67.10      & 84.04      & 64.48         & 86.37        \\
                                       & hotpotqa                 & 77.76       & 76.92      & 77.76         & 86.16        & 75.22        & 86.97       & 82.51      & 85.51      & 81.18         & 85.61        \\
                                       & MedQA                    & 67.41       & 59.74      & 67.41         & 82.31        & 64.98        & 82.31       & 77.32      & 82.25      & 76.52         & 81.43        \\
                                       & math                     & 78.05       & 82.01      & 78.05         & 87.14        & 79.67        & 87.75       & 88.81      & 89.11      & 90.37         & 89.60        \\
\midrule
\multirow{4}{*}{Qwen3-4B}              & 2Wiki          & 68.16       & 79.58      & 68.13         & 79.59        & 72.86        & 80.68       & 77.15      & 81.40      & 77.06         & 81.23        \\
                                       & hotpotqa                 & 67.86       & 86.07      & 67.86         & 86.08        & 63.38        & 86.44       & 83.23      & 86.13      & 84.19         & 87.02        \\
                                       & MedQA                    & 61.84       & 66.52      & 61.84         & 66.40        & 58.79        & 65.72       & 67.57      & 69.82      & 65.10         & 69.78        \\
                                       & math                     & 85.78       & 90.57      & 85.79         & 90.57        & 84.69        & 91.71       & 89.44      & 96.29      & 81.27         & 96.58        \\
\midrule
\multirow{4}{*}{Qwen3-14B}             & 2Wiki          & 78.29       & 82.57      & 78.29         & 82.57        & 78.36        & 83.05       & 72.11      & 82.18      & 73.50         & 83.03        \\
                                       & hotpotqa                 & 67.48       & 81.74      & 67.47         & 81.27        & 63.16        & 80.35       & 81.33      & 83.17      & 79.66         & 81.96        \\
                                       & MedQA                    & 70.23       & 76.67      & 70.23         & 77.32        & 66.38        & 74.86       & 67.00      & 70.57      & 66.12         & 70.18        \\
                                       & math                     & 83.00       & 95.06      & 83.02         & 95.04        & 88.01        & 95.30       & 92.32      & 98.68      & 90.01         & 98.76        \\
\midrule
\multirow{4}{*}{Qwen3-32B}             & 2Wiki          & 72.04       & 83.03      & 72.05         & 83.91        & 73.47        & 83.86       & 82.83      & 79.24      & 83.59         & 81.78        \\
                                       & hotpotqa                 & 68.74       & 87.78      & 68.75         & 87.62        & 66.23        & 87.31       & 84.23      & 84.70      & 84.20         & 85.40        \\
                                       & MedQA                    & 71.77       & 75.69      & 71.77         & 86.12        & 67.77        & 85.14       & 78.58      & 80.38      & 78.83         & 83.99        \\
                                       & math                     & 82.53       & 96.74      & 82.49         & 97.54        & 83.18        & 97.64       & 85.93      & 94.33      & 87.79         & 95.82        \\

\bottomrule
\end{tabular}
\end{table*}
\subsection{Human Annotation Study}
\label{app:human_annotation}
To assess the reliability of our LLM-based judging protocol, we perform a small-scale human verification study on outputs from two representative models (LLaMA-3.1-8B and Qwen3-4B) across three datasets (Math, HotpotQA, and TriviaQA). For each model--dataset combination, we randomly select 200 instances and manually evaluate one sampled response per instance. Human annotators provide binary correctness labels based solely on the semantic validity of the final answer. We then measure the agreement between human judgments and the automatic labels produced by our judge ensemble. As shown in Tab.~\ref{tab:human_annotation_subset}, the consistency remains high across all settings, indicating that the LLM-as-judge labeling is generally reliable.
\begin{table}[H]
\centering
\caption{Human annotation subsets by model size and dataset.}
\begin{tabular}{c|l|c}
\hline
\textbf{Model} & \textbf{Dataset} & \textbf{Consistency} \\
\hline
\multirow{3}{*}{LLama3.1-8B} & Math & 98\% \\
 & HotpotQA & 98\% \\
 & TriviaQA & 99\% \\
\hline
\multirow{3}{*}{Qwen3-4B} & Math & 99\% \\
 & HotpotQA & 98\% \\
 & TriviaQA & 99\% \\
\hline
\end{tabular}
\label{tab:human_annotation_subset}
\end{table}

\subsection{Different Candidate Ablation}
Consider a scenario where we sample 5 samples. Although the candidate answer has the highest conditional probability, it is not the one with the highest conditional probability in the largest semantic cluster in the semantic space. To produce a more reliable answer, we should consider candidates from the largest semantic cluster as our selection criterion. We have implemented this approach in all the results in the main experiment. The experimental results in Tab.~\ref{tab:candidate} demonstrate that our method still maintains strong robustness and stability.

\end{document}